\documentclass[11pt]{article}

\usepackage[final]{acl}

\usepackage{times}
\usepackage{latexsym}

\usepackage[T1]{fontenc}
\usepackage[utf8]{inputenc}

\usepackage{microtype}

\usepackage{inconsolata}

\usepackage{graphicx}

\usepackage{tikz}
\usepackage{amssymb}
\usepackage{pgfplots}
\pgfplotsset{compat=1.18}

\usepackage{amsmath}

\usepackage{booktabs}

\newcommand{\squishlist}{
 \begin{list}{$\bullet$}
  { \setlength{\itemsep}{0pt}
     \setlength{\parsep}{2pt}
     \setlength{\topsep}{2pt}
     \setlength{\partopsep}{0pt}
     \setlength{\leftmargin}{1em}
     \setlength{\labelwidth}{1em}
     \setlength{\labelsep}{0.4em} } }

\newcommand{\squishend}{
  \end{list}  }

\makeatletter
 \title{Balanced Accuracy: The Right Metric for Evaluating LLM Judges \\ Explained through Youden's J statistic%
 \ifacl@finalcopy
 \thanks{A preprint of this work is available on \href{https://arxiv.org/abs/2512.08121}{arXiv}.}%
 \fi
 }
\makeatother

\author{
 \textbf{Stephane Collot\textsuperscript{1}},
 \textbf{Colin Fraser\textsuperscript{2}},
 \textbf{Justin Zhao\textsuperscript{1}},
\\
 \textbf{William F. Shen\textsuperscript{1,3}},
 \textbf{Timon Willi\textsuperscript{1}},
 \textbf{Ilias Leontiadis\textsuperscript{1}},
\\
\\
 \textsuperscript{1}Meta Superintelligence Labs,
 \textsuperscript{2}Meta,
 \textsuperscript{3}University of Cambridge,
\\
 \small{
   \textbf{Correspondence:} \href{mailto:collot@meta.com}{collot@meta.com}
 }
}

\begin{document}
\maketitle
\begin{abstract}

Rigorous evaluation of large language models (LLMs) relies on comparing models by the prevalence of desirable or undesirable behaviors, such as task pass rates or policy violations. These prevalence estimates are produced by a classifier, either an LLM-as-a-judge or human annotators, making the choice of classifier central to trustworthy evaluation. Common metrics used for this choice, such as Accuracy, Precision, and F1, are sensitive to class imbalance and to arbitrary choices of positive class, and can favor judges that distort prevalence estimates. We show that Youden’s $J$ statistic is theoretically aligned with choosing the best judge to compare models, and that Balanced Accuracy is an equivalent linear transformation of $J$. Through both analytical arguments and empirical examples and simulations, we demonstrate how selecting judges using Balanced Accuracy leads to better, more robust classifier selection.

\end{abstract}

\section{Introduction}

Evaluating large language models (LLMs) is a cornerstone of their development cycle. Standard practice involves running models on benchmark datasets of user prompts and estimating the prevalence of key behaviors in their responses such as task pass rates, safety violations, or false refusals. Prevalence estimates rely on another classifier, typically an LLM, a fine-tuned model, or human annotators. We refer to this classifier as a \emph{judge} \cite{gu2024survey,liu-etal-2023-g,li2024llmsasjudges,li2024from,zheng2023judging}. Because prevalence measurements feed directly into ablation studies, capabilities assessments, and release decisions, the quality of this judge critically determines the validity of the resulting model comparisons.

However, despite widespread use of LLM-as-a-judge pipelines, there is less consensus on how to evaluate the judges themselves. We raise a central methodological question: \textit{Which metric best evaluates judges for the downstream task of comparing models on prevalence?}

In this position paper, we identify and advocate for a principled best practice grounded in the statistical structure of prevalence estimation, with a focus on judge-quality metrics measured on a golden set. We show that widely used metrics such as Accuracy, Precision, Recall, F1, and Macro-F1 are prevalence-dependent: they change as a function of the underlying class distribution, causing judges to be over- or under-valued depending on the dataset imbalance. As a result, these metrics less reliably reflect a judge’s ability to detect true differences between evaluated models.

We argue instead for \textbf{Balanced Accuracy} (equivalently, Macro-Recall) as the primary metric for judge selection. Balanced Accuracy is independent of class prevalence, assigns equal importance to both classes, extends naturally to multi-class settings, and most directly captures the key property needed for prevalence comparison: how well a judge distinguishes positive from negative instances. We formalize this by grounding the argument in Youden’s $J$ statistic \cite{youden1950index}, historically used in diagnostic testing to measure a classifier’s ability to separate classes. We show that $J$ is theoretically aligned with detecting prevalence differences and that Balanced Accuracy is a simple monotonic linear transformation of it. We provide geometric intuition through ROC analysis and empirical examples demonstrating that Balanced Accuracy leads to more reliable judge selection and more trustworthy downstream evaluation.

%\vspace{2mm}
\section{Preliminaries}
%\vspace{2mm}

This work focuses on \emph{pointwise} evaluation, where each model response is judged independently. This differs from \emph{pairwise} evaluation, where responses are compared head-to-head; pairwise settings produce inherently balanced labels, making metrics like Accuracy suitable for evaluating preference models \cite{malik2025rewardbench2advancingreward}.

We describe two datasets in our setup:
\squishlist
    \item [1. ] \textbf{Benchmark}: A dataset of prompts used to elicit responses from the evaluated LLMs, whose behavior prevalence we aim to compare.
    \item [2. ] \textbf{Golden set}: A dataset of prompts, responses, and gold labels used to evaluate the judges themselves.
\squishend

An ideal golden set is balanced across classes to enable precise measurement of judge performance. In practice, this is difficult to obtain: ground-truth labels are unknown during dataset construction; rare behaviors (e.g., safety violations) are costly to collect; and a high-quality set must include responses from multiple models to capture model-specific biases. Downsampling wastes expensive gold labels, while upsampling introduces artificial distribution shifts. Consequently, golden sets are typically imbalanced, underscoring the need for judge metrics, such as Balanced Accuracy, that remain valid under class imbalance.

\section{The Pitfalls of Traditional Metrics}

When comparing multiple judges, we need a single, principled metric that reflects how well each judge will support the downstream task of estimating behavior prevalence across LLMs. A suitable judge metric should satisfy three core criteria:

\squishlist
% \begin{enumerate}
    \item[1. ] \textbf{Prevalence independence}: It should not change when the class distribution of the golden set changes.
    \item[2. ] \textbf{Label symmetry}: Flipping which class is designated “positive” should not alter the metric’s meaning.
    \item[3. ] \textbf{Balanced class treatment}: It should capture a judge’s ability to correctly identify both positive and negative instances, since both directly affect prevalence estimation.
% \end{enumerate}
\squishend

This section outlines the key issues with commonly used metrics.

\subsection{Precision and Recall vs. Sensitivity and Specificity}

Precision and Recall are widely used for evaluating binary classifiers, but they have structural properties that make them unsuited for judge selection.

\textbf{Lack of label symmetry.} Precision and Recall treat the “positive” class as privileged. When we flip class labels—for example, defining “safe” instead of “violating” as the positive class—Recall simply becomes Recall for the other class, but Precision does \emph{not}: it turns into Negative Predictive Value (NPV). This asymmetry creates inconsistencies across datasets or benchmarks that use different labeling conventions. In contrast, Sensitivity and Specificity (true positive rate and true negative rate) form a \emph{label-symmetric} pair: swapping class labels simply swaps the two metrics. This makes them a more principled basis for judge evaluation.

\textbf{Prevalence dependence.} Precision is dependent on class prevalence because it conditions on predicted positives. When positives are rare, even a few false positives can drastically reduce Precision, regardless of how well the judge performs on negatives. As golden sets are typically unbalanced and expensive to construct, a metric sensitive to prevalence is undesirable. Sensitivity and Specificity measure conditional accuracy within each class and therefore remain stable across datasets with different class ratios.

\textbf{Why not keep a pair of metrics?} Although Sensitivity and Specificity form a robust and prevalence-independent pair, comparing judges using two numbers invites ambiguity: one judge may have higher Sensitivity while another has higher Specificity. This motivates a \emph{single} summary statistic such as Youden’s $J$ that combines them without sacrificing the desirable properties of symmetry and prevalence independence.

The same reasoning applies to AUC metrics: PR-AUC inherits Precision's prevalence dependence, while ROC-AUC — built from Sensitivity and Specificity — is prevalence-independent and label-symmetric, making it more relevant for our task. PR-based summaries can also exhibit high statistical uncertainty under small test sets, particularly in the low-recall regime where the joint uncertainty becomes highly non-linear \cite{urlus2023pointwise}.

\subsection{Issues with the F1 Score}

The F1 score (also called binary F1 score, or micro average F1 score in binary classification) combines Precision and Recall into a single metric, but it inherits the same prevalence dependence and label asymmetry issues as its constituent metrics.

% \vspace{-4mm}

\begin{small}
\begin{align*}
F_1 &= \frac{2}{\text{recall}^{-1} + \text{precision}^{-1}} \\ & = 2 \frac{\text{precision} \cdot \text{recall}}{\text{precision} + \text{recall}} \\
&= \frac{2\text{TP}}{2\text{TP} + \text{FP} + \text{FN}}
\end{align*}
\end{small}

Most critically, F1 completely ignores True Negatives (TNs), depending only on TP, FP, and FN. A judge that performs poorly on negatives but excellent on positives can still achieve a high F1 score, even though prevalence estimation requires balanced performance on both classes. For tasks where both false positives and false negatives directly influence model comparison, ignoring TNs is a fatal flaw.

\subsection{Issues with the Macro-Averaged F1 Score}

Macro-F1 averages the F1 scores of the positive and negative classes and is therefore label symmetric:
%\vspace{-1mm}
\begin{small}
$$ \mathrm{Macro\text{-}F1} = \tfrac{1}{2}\,(F1_\text{positive} + F1_\text{negative}). $$
\[
\begin{aligned}
\mathrm{Macro\text{-}F1} &= \frac{\mathrm{TP}}{2\,\mathrm{TP} + \mathrm{FP} + \mathrm{FN}} \\
&\quad + \frac{\mathrm{TN}}{2\,\mathrm{TN} + \mathrm{FP} + \mathrm{FN}}
\end{aligned}
\]
\end{small}

However, it retains a deeper problem: each class-specific F1 score is still prevalence dependent, and its value is highly dependent on the class prevalence of the golden set.

For example, the $F1_{\text{positive}}$ score is the harmonic mean of prevalence-independent $\mathrm{Recall_{\text{positive}}} = \frac{\mathrm{TP}}{\mathrm{TP} + \mathrm{FN}}$ and highly prevalence-dependent $\mathrm{Precision_{\text{positive}}} = \frac{\mathrm{TP}}{\mathrm{TP} + \mathrm{FP}}$.

When the positive class is rare, $\mathrm{Precision_{\text{positive}}}$ is especially unstable. Even a handful of errors on the majority class can drastically change Macro-F1, making it less reliable across golden sets with different class ratios. This instability is problematic for judge selection, where we want a metric that remains meaningful across datasets constructed from different models or sampling distributions.

\subsection{Issues with Accuracy}

Accuracy, defined as $\frac{\text{TP} + \text{TN}}{\text{TP} + \text{TN} + \text{FP} + \text{FN}}$, is perhaps the most widely used metric. In heavily imbalanced golden sets, which are common in safety evaluations, for example, a classifier can achieve high Accuracy simply by predicting the majority class. This makes Accuracy ill-suited for our goal of selecting judges who perform well across both classes \cite{dorner2025limits}.

\subsection{Issues with Agreement Metrics}

Agreement measures such as Cohen's kappa \cite{cohen1960coefficient}, Scott's Pi \cite{scott1955reliability}, and Krippendorff's alpha \cite{krippendorff2004content} are frequently used in human annotation settings. Similarly, correlation-based metrics such as Pearson correlation are used to measure alignment between judge scores and human ratings in pointwise rubric-based evaluation \cite{kim2024prometheus}. However, these metrics measure the wrong quantity for our purposes.

Agreement metrics quantify \emph{inter-rater reliability}: the degree to which two annotators label items consistently, adjusting for chance agreement. But for judge selection, our goal is \emph{accuracy against ground truth}, not consistency with another label source. Correcting for chance is irrelevant when the ground truth is known.

Agreement metrics also suffer from prevalence dependence. When one class is rare, their chance-correction terms can cause the metric to collapse toward zero, even when a classifier performs well on the minority class.

\section{Youden’s J: A Metric Theoretically Aligned}

The shortcomings of traditional metrics suggest that we should instead evaluate judges using a measure that (i) is independent of class prevalence, (ii) treats both classes symmetrically, and (iii) reflects how well the judge preserves true differences in prevalence between models. Youden’s $J$ statistic \cite{powers2011evaluation} satisfies all three criteria and emerges naturally from the structure of the prevalence-estimation problem. Youden’s $J$ is defined for any binary classifier as:

%\vspace{-5mm}
\begin{small}
$$ J = \text{Sensitivity} + \text{Specificity} - 1 $$
\end{small}
where Sensitivity (TPR) captures performance on the positive class and Specificity (TNR) captures performance on the negative class. Because both TPR and TNR are conditional measures, they are unaffected by the underlying class distribution, making $J$ prevalence independent and symmetric under class-label swaps. This expression can be written equivalently as:

%\vspace{-5mm}
\begin{small}
$$ J = \text{Positive Recall} + \text{Negative Recall} - 1 $$
$$ J = \text{TPR} + \text{TNR} - 1 $$
$$ J = \text{TPR} - \text{FPR} $$
$$ J = \frac{TP \times TN - FP \times FN}{(TP + FN)(TN + FP)} $$
\end{small}

Its alternative names include ``Net Detection Rate'' (as the formula $\text{TPR} - \text{FPR}$ can be interpreted as the rate of true detections net of false detections), ``Informedness'', ``Bookmaker Informedness'', and ``$\Delta P'$''.

The value of $J$ ranges from $-1$ to $1$, with $0$ corresponding to random guessing, positive values indicating better-than-chance classification, and negative values indicating systematic misclassification (which could be corrected by flipping the classifier’s output).

\subsection{The Natural Emergence of Youden’s J: The Classifier Slope}

To understand how $J$ naturally emerges from our evaluation goal, consider how a judge distorts prevalence estimates. Let the true prevalence of a behavior be $x$, and let $y$ be the prevalence measured by a judge with true-positive rate TPR and false-positive rate FPR. Under pointwise classification, the measured prevalence is:

\begin{small}
$$ y = \text{TPR} \cdot x + \text{FPR} \cdot (1-x) $$
\end{small}

Now, consider two models with a true prevalence difference of $\Delta x$. The judge will measure a difference of:
\begin{small}
$$ \Delta y = (\text{TPR} - \text{FPR}) \cdot \Delta x $$
\end{small}

\begin{figure}[h!]
\centering
\begin{tikzpicture}

% --- Define parameters for the plot ---
% We can choose values that look like the plot.
% Let's set \beta = 0.2 and \alpha = 0.7 for a shallower slope
\pgfmathsetmacro{\betaVal}{0.2}
\pgfmathsetmacro{\alphaVal}{0.7}
% Calculate the intersection point x'
% x' = \beta + (\alpha - \beta)x'
% x'(1 - (\alpha - \beta)) = \beta
% x' = \beta / (1 - \alpha + \beta)
\pgfmathsetmacro{\muPrime}{ \betaVal / (1 - \alphaVal + \betaVal) } % This calculates to 0.5

\begin{axis}[
    % --- Axes Labels ---
    xlabel={Actual Prevalence $x$},
    ylabel={Estimated prevalence $y$},
    ylabel style={align=center}, % Center-align label
    % --- Axes Limits and Lines ---
    xmin=0, xmax=1,
    ymin=0, ymax=1,
    enlarge x limits={upper=0.1}, % Add 10% space to the right
    enlarge y limits={upper=0.1}, % Add 10% space to the top
    axis x line=bottom, % x-axis line at bottom
    axis y line=left,   % y-axis line at left
    axis line style={-stealth}, % Add arrows to the end of axes
    % --- Ticks ---
    xtick={0, \muPrime, 1},
    xticklabels={0, $x'$, 1},
    ytick={0, \betaVal, \alphaVal, 1},
    yticklabels={0, $\beta$, $\alpha$, 1},
    % --- General Style ---
    width=0.98\columnwidth, % Set the size of the plot
    height=0.98\columnwidth,
    clip=false, % Allow drawing/nodes outside the axis box
    every tick label/.style={font=\small},
    label style={font=\small},
    title style={yshift=5pt},
]

% --- Plot 1: Perfect Labeler (y=x) ---
\addplot[
    domain=0:1,
    samples=2, % Only need 2 points for a straight line
    color=blue!70!white, % A slightly lighter blue
    dashed,
    thick,
] {x}
% Label for the blue line
node[blue!70!white, font=\small, align=left, pos=0.70, sloped, above right, xshift=-18mm, yshift=0mm] 
    {perfect labeler \\ $y = x$};
% --- Plot 2: Imperfect Labeler (y = \beta + (\alpha-\beta)x) ---
\addplot[
    domain=0:1,
    samples=2,
    color=red,
    thick,
] {\betaVal + (\alphaVal - \betaVal) * x}
% Label for the red line
node[red, font=\small, pos=0.25, sloped, above, yshift=-1mm] 
    {$y\!=\!\beta\!+\!(\alpha\!-\!\beta)x$};
% --- Dotted Lines ---
% Vertical line from (x', 0) to intersection point (x', x')
\draw[dotted, thick] (axis cs: \muPrime, 0) -- (axis cs: \muPrime, \muPrime);
% Vertical line on right axis
\draw[dotted, thick] (axis cs: 1, 0) -- (axis cs: 1, 1);
% Manual tick marks on right axis
\draw[thin] (axis cs: 1.02, 0) -- (axis cs: 1, 0); % tick at 0
\draw[thin] (axis cs: 1.02, \alphaVal) -- (axis cs: 1, \alphaVal); % tick at alpha
\draw[thin] (axis cs: 1.02, 1) -- (axis cs: 1, 1); % tick at 1
% Alpha label on right side
\node[font=\small, black] at (axis cs: 1.01, \alphaVal) [right] {$\alpha$};
% --- Bias Arrows & Labels ---
% Positive Bias (at x=0)
\def\biasXshift{0.03} % Shift slightly right from y-axis
\pgfmathsetmacro{\yMidPos}{ \betaVal / 2 } % Midpoint for the label
\node[font=\small, align=left, black] at (axis cs: \biasXshift+0.02, \yMidPos) [right] 
    {positive bias \\ overestimate};
% Continuous arrow with arrows at both ends
\draw[red, <->, thick] (axis cs: \biasXshift, 0.01) -- (axis cs: \biasXshift, \betaVal - 0.01);

% Negative Bias (at x=1)
\def\biasXshiftNeg{0.97} % Shift slightly left from x=1
\pgfmathsetmacro{\yMidNeg}{ (\alphaVal + 1) / 2 } % Midpoint for the label
\node[font=\small, align=right, black] at (axis cs: 0.98, \yMidNeg + 0.02) [left] 
    {negative bias \\ underestimate};
% Continuous arrow with arrows at both ends
\draw[red, <->, thick] (axis cs: \biasXshiftNeg, \alphaVal + 0.01) -- (axis cs: \biasXshiftNeg, 1 - 0.01);

% --- Delta Y / Delta X Triangle ---
\def\deltaXval{0.25} % Define the width of the triangle
\pgfmathsetmacro{\xStart}{\muPrime} % Start at intersection x
\pgfmathsetmacro{\yStart}{\muPrime} % Start at intersection y
\pgfmathsetmacro{\xEnd}{\xStart + \deltaXval} % End x
\pgfmathsetmacro{\yEnd}{\betaVal + (\alphaVal - \betaVal) * \xEnd} % End y (on the red line)

% Draw horizontal line \Delta x
\draw[green!50!black, ->, thick] (axis cs: \xStart, \yStart) -- (axis cs: \xEnd, \yStart)
    node[midway, below, black, font=\small] {$\Delta x$};
    
% Draw vertical line \Delta y
\draw[green!50!black, ->, thick] (axis cs: \xEnd, \yStart) -- (axis cs: \xEnd, \yEnd)
    node[midway, right, black, font=\small] {$\Delta y$};
    
\end{axis}
\end{tikzpicture}
\caption{Illustration of how an imperfect classifier affects prevalence estimation. The red line shows the relationship $y\!=\!\beta\!+\!(\alpha\!-\!\beta)x$ between actual prevalence $x$ and estimated prevalence $y$ for a classifier with sensitivity $\alpha$ and specificity $(1\!-\!\beta)$. The slope is $(\alpha\!-\!\beta)\!=\!J$, showing that Youden’s $J$ directly measures the classifier's ability to preserve true prevalence differences.}
\label{fig:classifier_slope}
\end{figure}
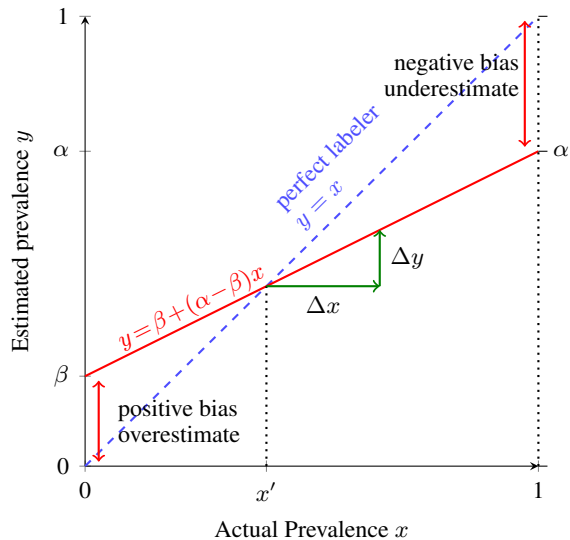

This expression shows that the judge acts as a \textit{linear filter}: it preserves the true difference but scales it by a factor of $(\text{TPR} - \text{FPR})$, which is exactly Youden's $J$. A judge with a higher $J$ more faithfully preserves the magnitude of true prevalence differences and produces stronger, more reliable signals when comparing models. This relationship is illustrated in Figure \ref{fig:classifier_slope}, which shows how an imperfect classifier affects prevalence estimation.

%\vspace{-5pt}

\subsection{Relationship to ROC Analysis}

\begin{figure}[h!]
\centering
\begin{tikzpicture}
\begin{axis}[
    title={ROC Chart},
    xlabel={\shortstack{False Positive Rate\\1 - Specificity}},
    ylabel={\shortstack{True Positive Rate\\Sensitivity}},
    xmin=0, xmax=1,
    ymin=0, ymax=1,
    xtick={0,0.2,0.4,0.6,0.8,1.0},
    ytick={0,0.2,0.4,0.6,0.8,1.0},
    legend pos=south east,
    width=0.95\columnwidth,
    height=0.9\columnwidth,
    grid=none,
    axis lines=left,
    every tick label/.style={font=\footnotesize},
    label style={font=\footnotesize},
    title style={font=\footnotesize},
]

% ROC Curve
\addplot[red, very thick] coordinates {
    (0.0, 0.0)
    (0.08, 0.62)
    (0.14, 0.64)
    (0.25, 0.90)
    (0.25, 0.95)
    (0.45, 0.97)
    (0.85, 0.97)
    (0.85, 1.0)
    (1.0, 1.0)
};

% Diagonal reference line
\addplot[dashed, domain=0:1, black]{x};

% Youden’s $J$ index
\draw[black, <->, thick] (axis cs:0.25, 0.95) -- (axis cs:0.25, 0.25);
\node[font=\footnotesize] at (axis cs:0.32, 0.6) {J};

\end{axis}
\end{tikzpicture}
\caption{The relationship between Youden’s $J$ and the ROC curve. J is the vertical distance between a point on the ROC curve for a given threshold and the diagonal chance line. The optimal threshold needs to balance sensitivity and specificity, and is represented by the point on the curve that maximizes this vertical distance.}
\label{fig:roc_youden}
\end{figure}
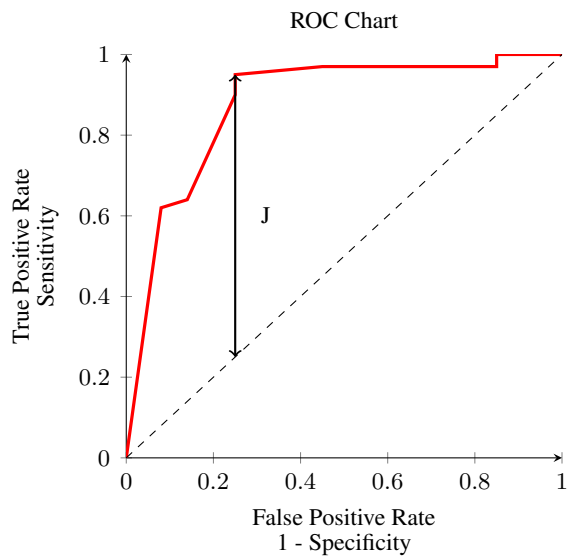

Youden’s $J$ also has a clear geometric interpretation in ROC space \cite{fawcett2006introduction}. Each threshold of a classifier corresponds to a point (FPR, TPR) on the ROC curve, and $J$ is the vertical distance between this point and the diagonal chance line ($y=x$). (Figure \ref{fig:roc_youden}).

A key practical advantage of Youden's $J$ over ROC AUC is that it does not require predicted probabilities from the judge; it can be computed directly from binary labels. This makes it equally applicable to LLM judges (where logits might be available) and single-review human annotations (where they are not).
%\vspace{+15pt}

%\vspace{-5pt}

\textbf{Remarks on calibration:} When calibrating the decision threshold, we should select the threshold that maximizes Youden’s $J$. This requires labeled data to tune the threshold in addition to the golden set to prevent overfitting. In this paper we estimate prevalence by taking the mean of binary labels. Alternatively, one could estimate prevalence by taking the mean of the predicted score (a less common practice), in which case a metric like the Kuiper statistic \cite{kuiper1960tests}, which assesses calibration across the entire range of predicted probabilities, is relevant.

In empirical analyses on our datasets, we observe that $J$ is highly correlated with ROC AUC (0.88), consistent with their shared reliance on TPR and FPR. By contrast, the correlation between $J$ and F1 is substantially weaker (0.56), reflecting F1's dependence on Precision and hence on class prevalence. These observations support the theoretical distinction between prevalence-independent and prevalence-dependent metrics. More broadly, recent work has shown that the geometry of the ROC curve also determines the scaling behavior of test-time methods such as Best-of-N and rejection sampling when using LLM judges as verifiers \cite{dorner2025roc}.

\section{Balanced Accuracy: A Practical and Aligned Metric}

While Youden’s $J$ provides the theoretical foundation for selecting an optimal judge, we recommend using \textit{Balanced Accuracy} as the primary metric for reporting and comparing judge quality in practice. Balanced Accuracy captures exactly the same information as $J$, but presents it on the familiar and intuitive $0-1$ scale, making it easier to communicate and interpret in evaluation contexts.

Balanced Accuracy is defined as the arithmetic mean of Sensitivity and Specificity, or equivalently of positive and negative recall:

%\vspace{-4mm}
\begin{small}
$$ \text{Balanced Accuracy} = \frac{\text{Sensitivity} + \text{Specificity}}{2} $$
$$ \text{Balanced Accuracy} = \frac{\text{Recall}_\text{pos} + \text{Recall}_\text{neg}}{2} $$
\end{small}
%\vspace{1mm}
Balanced Accuracy is linearly related to Youden’s 
$J$:

%\vspace{-2mm}
\begin{small}
$$ \text{Balanced Accuracy} = \frac{\text{Youden’s $J$} + 1}{2} $$
$$ \text{Youden’s $J$} = 2 \times \text{Balanced Accuracy} - 1 $$
\end{small}

Because this transformation is monotonic, Balanced Accuracy and $J$ rank judges identically and share all the theoretical advantages previously established for $J$: independence from class prevalence, symmetry under class label swapping, and alignment with our goal of detecting true differences in prevalence between models.

From a practical standpoint, Balanced Accuracy benefits from being easy to interpret, bounded in 
$[0, 1]$, and supported in common libraries (e.g., scikit-learn). Its similarity to standard accuracy makes it easy to communicate. In the case of a perfectly balanced golden set, Balanced Accuracy and Accuracy are equal.

\subsection{Multi-Class Generalization}

Balanced Accuracy extends naturally to multi-class classification by averaging recall uniformly across all classes:

%\vspace{-1mm}
\begin{small}
$$ \text{Balanced Accuracy} = \frac{1}{n} \sum_{i=1}^{n} \text{Recall}_i $$
\end{small}

This generalization preserves the key properties from the binary case: each class is treated symmetrically and classes are weighted equally, which avoids distortions caused by class imbalance. Balanced Accuracy remains appropriate for evaluating multi-class judges when the task involves comparing prevalence across multiple behavior categories or their ability to detect relative prevalence differences.

Youden’s $J$ can also be extended to multi-class settings—typically by computing a one-vs-rest 
$J$ value for each class and taking the macro-average (Macro $J$). Balanced Accuracy and Macro $J$ remain linearly related in the multi-class case. For example, for $n$ classes:

%\vspace{-4mm}
\begin{small}
$$\text{Balanced Accuracy} = \left(\frac{n-1}{n}\right)\! \times\! \text{Macro J} + \left(\frac{1}{n}\right)$$
\end{small}
%\vspace{25mm}
\section{Empirical Studies}

We present three empirical studies to illustrate how the choice of metric directly affects judge selection and downstream evaluation quality. First, we analyze two real-world scenarios—one involving safety violation detection and another involving response-format compliance—where two candidate judges are compared on their respective golden sets. Although the underlying datasets are proprietary, the intuitions do not depend on the specifics of these benchmarks. Finally, we present a large-scale simulation study that systematically quantifies how metric choice influences a judge’s ability to correctly rank candidate assistant models by prevalence.

\subsection{Case 1: Precision, Recall, and Specificity trade-offs}
Two judges are evaluated on a golden set with a violation prevalence of 8.3\%. Judge A has much higher Recall, while Judge B has better Precision and Specificity, this is a common trade-off. As shown in Table \ref{tab:case1}, F1, Macro-F1, and Accuracy all favor Judge B. However, Balanced Accuracy correctly identifies Judge A as superior due to its better balance of TPR and FPR.

\begin{table}[h!]
% \begin{table}[t]
\raggedright
\small
\setlength{\tabcolsep}{2pt}
\begin{tabular}{lcccc}
\toprule
\textbf{Judge} & \textbf{Precision} & \textbf{Recall} & \textbf{Specificity} & \textbf{Youden’s J} \\
\midrule
Judge A & 0.32 & \textbf{0.76} & 0.85 & \textbf{0.61} \\
Judge B & \textbf{0.41} & 0.57 & \textbf{0.92} & 0.49 \\
\bottomrule
\end{tabular}
\hspace{0.5em}
\begin{tabular}{lccccc}
\toprule
\textbf{Judge} & \textbf{F1} & \textbf{Macro-F1} & \textbf{Accuracy} & \textbf{Balanced Accuracy} \\
\midrule
Judge A & 0.45 & 0.68 & 0.85 & \textbf{0.81} \\
Judge B & \textbf{0.47} & \textbf{0.71} & \textbf{0.90} & 0.75 \\
\bottomrule
\end{tabular}
\hspace{0.5em}
\begin{tabular}{lccc}
\toprule
\textbf{Judge} & \textbf{Prevalence} & \textbf{TP/FP/TN/FN} \\
\midrule
Judge A & 0.083 & 63/133/784/20 \\
Judge B & 0.083 & 47/69/848/36 \\
\bottomrule
\end{tabular}
\caption{Case 1: Performance metrics for two judges evaluated on a golden set with 8.3\% prevalence (1000 samples). While Balanced Accuracy correctly identifies Judge A as better, the metrics F1, Macro-F1, and Accuracy would have picked the wrong Judge B.}
\label{tab:case1}
\end{table}
%\vspace{-5pt}

\subsection{Case 2: Perfect Precision and Specificity vs. Balanced Performance}
Two judges are evaluated on a golden set with 20\% prevalence. As shown in Table \ref{tab:case2}, Judge B has perfect Precision (1.00) and Specificity (1.00) but lower Recall (0.20), making it overly conservative and missing many true positives. Judge A has higher Recall (0.25) but lower Specificity (0.975) and lower Precision (0.71), resulting in more balanced performance. Balanced Accuracy captures this difference, identifying Judge A as better, while Accuracy favors Judge B due to its perfect specificity and low positive class prevalence.

\begin{table}[h!]
\setlength{\tabcolsep}{2pt}
\raggedright
\small
% \newcommand{\tablewidth}{0.95\textwidth}  % Define a common width
% \begin{tabular}{\tablewidth}{lcccc}
\begin{tabular}{lcccc}
\toprule
\textbf{Judge} & \textbf{Precision} & \textbf{Recall} & \textbf{Specificity} & \textbf{Youden’s J} \\
\midrule
Judge A & 0.71 & \textbf{0.25} & 0.975 & \textbf{0.225} \\
Judge B & \textbf{1.00} & 0.20 & \textbf{1.00} & 0.20 \\
\bottomrule
\end{tabular}
\hspace{0.5em}
% \begin{tabular}{\tablewidth}{lccccc}
\begin{tabular}{lccccc}
\toprule
\textbf{Judge} & \textbf{F1} & \textbf{Macro-F1} & \textbf{Accuracy} & \textbf{Balanced Accuracy} \\
\midrule
Judge A & \textbf{0.37} & \textbf{0.64} & 0.83 & \textbf{0.61} \\
Judge B & 0.33 & 0.62 & \textbf{0.84} & 0.60 \\
\bottomrule
\end{tabular}
\hspace{0.5em}
% \begin{tabular}{\tablewidth}{lccc}
\begin{tabular}{lccc}
\toprule
\textbf{Judge} & \textbf{Prevalence} & \textbf{TP/FP/TN/FN} \\
\midrule
Judge A & 0.20 & 50/20/780/150 \\
Judge B & 0.20 & 40/0/800/160 \\
\bottomrule
\end{tabular}
\caption{Case 2: Performance metrics for two judges evaluated on a golden set with 20\% prevalence (1000 samples). While Balanced Accuracy correctly identifies Judge A as better, the metric Accuracy would have picked the wrong Judge B.}
\label{tab:case2}
\end{table}
%\vspace{-5pt}

\subsection{Simulated Judge Selection for Model Ranking}

To evaluate how metric choice affects downstream ranking of candidate assistant models, we simulate 100{,}000 scenarios in a Monte-Carlo fashion. In each scenario, we generate three candidate judges with $(\mathrm{TPR},\mathrm{FPR})$ sampled from $\mathrm{Uniform}(0, 1)$, and five assistant models with true violation prevalences sampled from $\mathrm{Uniform}(0.01, 0.5)$. For each judge, we measure its
\emph{true} quality by applying it to 200 benchmark samples per model and
computing its pairwise model-ranking accuracy (RankAcc). Separately, each
judge is evaluated on a golden set of 800 labeled examples, from which we
compute four scalar metrics: Balanced Accuracy, Accuracy, F1, and Macro-F1, and select the top-scoring judge under each metric.

Table~\ref{tab:case3} reports two outcomes: (i) the probability that each metric selects the RankAcc-best judge and (ii) the average degradation in RankAcc when it does not. Balanced Accuracy attains the highest success rate ($75.2\%$) and the smallest ranking-accuracy loss ($0.033$), substantially outperforming Accuracy ($67.5\%$), F1 ($61.7\%$), and Macro-F1 ($70.7\%$). Selecting judges with Accuracy or F1 produces roughly 30--50\% more ranking error compared to selecting with Balanced Accuracy.

\begin{table}[h!]
\centering
\small
\begin{tabular}{lcc}
\toprule
\textbf{Metric} & \textbf{Success Rate} & \textbf{Avg Rank Gap} \\
\midrule
Balanced Accuracy & \textbf{0.752} & \textbf{0.033} \\
Macro-F1 & 0.707 & 0.049 \\
Accuracy & 0.675 & 0.067 \\
F1 & 0.617 & 0.094 \\
\bottomrule
\end{tabular}
\caption{Ranking simulation results across 100{,}000 scenarios with 3 judges and 5 models. Success rate measures how often each metric selected the rank-optimal judge. Avg rank gap measures the average loss in ranking accuracy when the metric's selected judge differs from the rank-optimal judge.}
\label{tab:case3}
\end{table}
%\vspace{-5pt}

In ablations on golden set prevalence, golden set size, and model evaluation size, we find that Balanced Accuracy is consistently the scalar metric most aligned with the practical goal of choosing judges that preserve the true ordering of assistant models of varying prevalence (Appendix \ref{app:simulation_details}).

\section{Conclusion}

In the industrial setting of evaluating large language models, selecting appropriate metrics for assessing judges is essential for producing trustworthy prevalence estimates and, ultimately, for making good model development and release decisions. In this paper, we have shown that Balanced Accuracy offers a principled and practical choice for benchmarking judges. It captures the same theoretical foundations as Youden’s $J$ including prevalence independence, symmetry under class label definitions, and balanced treatment of errors, all while presenting results on a familiar $0-1$ scale that is easier to interpret and communicate. 

Our empirical studies demonstrate that relying on commonly used metrics such as Accuracy or F1 can misrepresent judge quality, particularly in imbalanced settings, and can lead to selecting judges that distort true differences in model behavior. Balanced Accuracy, by contrast, consistently preserves these differences and supports more reliable downstream comparisons.

We hope that adopting Balanced Accuracy as a standard metric for judge evaluation will contribute to more robust assessment practices for LLMs.

% EACL requires a MANDATORY "Limitations" section just before the references.
% Failure to include this section will result in desk rejection.
\section*{Limitations}

While we recommend Balanced Accuracy as a principled and practical primary metric for evaluating judges, it is not without limitations.

\textbf{First, Balanced Accuracy summarizes performance through TPR and TNR, but it does not fully characterize a judge’s error profile.} A judge with a higher Balanced Accuracy may still have lower precision, higher false-positive rates, or other operational characteristics that matter for specific applications (e.g., safety teams may prioritize very high recall on violation classes), while statistical debiasing approaches require high correlation between judge and ground truth scores to provide meaningful sample efficiency gains \cite{dorner2024limits}. As with any scalar metric, Balanced Accuracy should be complemented with inspection of the underlying confusion matrix and class-specific error rates.

\textbf{Second, our analysis assumes that a judge’s error rates (TPR and FPR) remain stable across the assistant models being compared.} If a judge exhibits model-specific bias such as the self-preference and family-preference effects observed when LLM judges evaluate outputs from related models \cite{Wataoka2024SelfPreference,zheng2023judging,dorner2025limits,chen2024humans}, then prevalence estimates may be distorted regardless of the metric used. Detecting and mitigating this form of judge bias is an essential but orthogonal challenge that must accompany any metric-based judge selection.

\textbf{Third, in multi-class settings, Balanced Accuracy weights each class equally, which may be inappropriate for tasks with long-tailed label distributions.} Rare classes with sparse or noisy annotations can disproportionately influence the metric, making the overall score more volatile and potentially misaligned with evaluation priorities. In such settings, practitioners may prefer variants that reweight classes.

Finally, \textbf{Balanced Accuracy does not eliminate the need for task-specific judgment.} For high-stakes domains, qualitative inspection, secondary metrics (e.g., raw TPR/FPR), and bias analyses remain essential for judge selection. Balanced Accuracy provides a strong and reliable foundation, but it is best viewed as part of a broader evaluation toolkit.

\bibliography{custom}

\newpage
\appendix

\section{Ranking Accuracy Simulation Details}
\label{app:simulation_details}

\subsection{Simulation procedure}
To compare different scalar metrics for judge selection, we construct a synthetic
evaluation environment in which we can directly observe each judge's true downstream
utility and how reliably each metric identifies that judge from finite data. 
Each simulation scenario proceeds as follows.

\textbf{Judges and models.}
We sample $K$ assistant models, each with a ground-truth violation prevalence
$p_k \sim \mathrm{Uniform}(0.01, 0.5)$ (sorted so that the true ranking is defined).
We also sample $J$ candidate judges, each parameterized by a sensitivity--specificity
pair $(\mathrm{TPR}_j, \mathrm{FPR}_j)$ drawn independently from $\mathrm{Uniform}(0,1)$.
A judge therefore outputs ``violation'' on a model-$k$ example with probability:
\[
q_{jk} = p_k \cdot \mathrm{TPR}_j + (1 - p_k)\cdot \mathrm{FPR}_j
\]

\textbf{True downstream ranking quality.}
Each judge $j$ is applied to $n_{\text{eval}}$ independent samples from each model,
producing estimated violation rates $\hat{p}_{jk}$. 
We compute the judge’s \emph{pairwise model-ranking accuracy} (RankAcc):

\begin{small}
\[
\mathrm{RankAcc}_j = 
\frac{\#\{(a,b) : \mathrm{sign}(p_b - p_a) = \mathrm{sign}(\hat{p}_{jb} - \hat{p}_{ja})\}}
     {\#\text{model pairs}}
\]
\end{small}

The judge with the highest RankAcc is taken to be the \emph{ground-truth best judge} for the scenario.

\textbf{Golden-set evaluation and metric-based selection.}
Independently of model evaluation, we simulate a golden set of size 
$n_{\text{golden}}$ with prevalence $p_{\text{golden}}$. 
We draw $n_{\text{pos}} \sim \mathrm{Binomial}(n_{\text{golden}}, p_{\text{golden}})$ 
and $n_{\text{neg}} = n_{\text{golden}} - n_{\text{pos}}$.
For each judge, we sample the corresponding TP/FP/FN/TN counts using its
$(\mathrm{TPR}_j, \mathrm{FPR}_j)$, and compute four scalar metrics:
Balanced Accuracy, Accuracy, positive-class F1, and Macro-F1.
For a given metric $m$, we select the judge with the highest score.

\textbf{Outcome measures.}
Across $N$ simulated scenarios (typically 10k--20k), we evaluate each metric $m$ using:

\vspace{3mm}

\textbf{1. Selection correctness:}
\begin{small}
\small\[
\Pr\!\left(m \text{ selects the same judge as } \arg\max_j \mathrm{RankAcc}_j\right)
\]
\end{small}

\textbf{2. Expected ranking loss:}
\[
\mathbb{E}\!\left[
    \mathrm{RankAcc}_{\text{best}} 
    - \mathrm{RankAcc}_{\text{chosen by } m}
\right]
\]

These quantities measure how often metric-based selection recovers the
rank-optimal judge and how much downstream ranking quality is lost when it does not.

\subsection{Results}

Over 100k scenarios (with default settings $K=5$ models, $J=3$ judges, $n_{\text{eval}} = 2{,}000$ model-eval samples per model and $n_{\text{golden}} = 2{,}000$), we obtain Table \ref{tab:case3} in the main paper.

\begin{figure}[t]
    \centering
    \includegraphics[width=\linewidth]{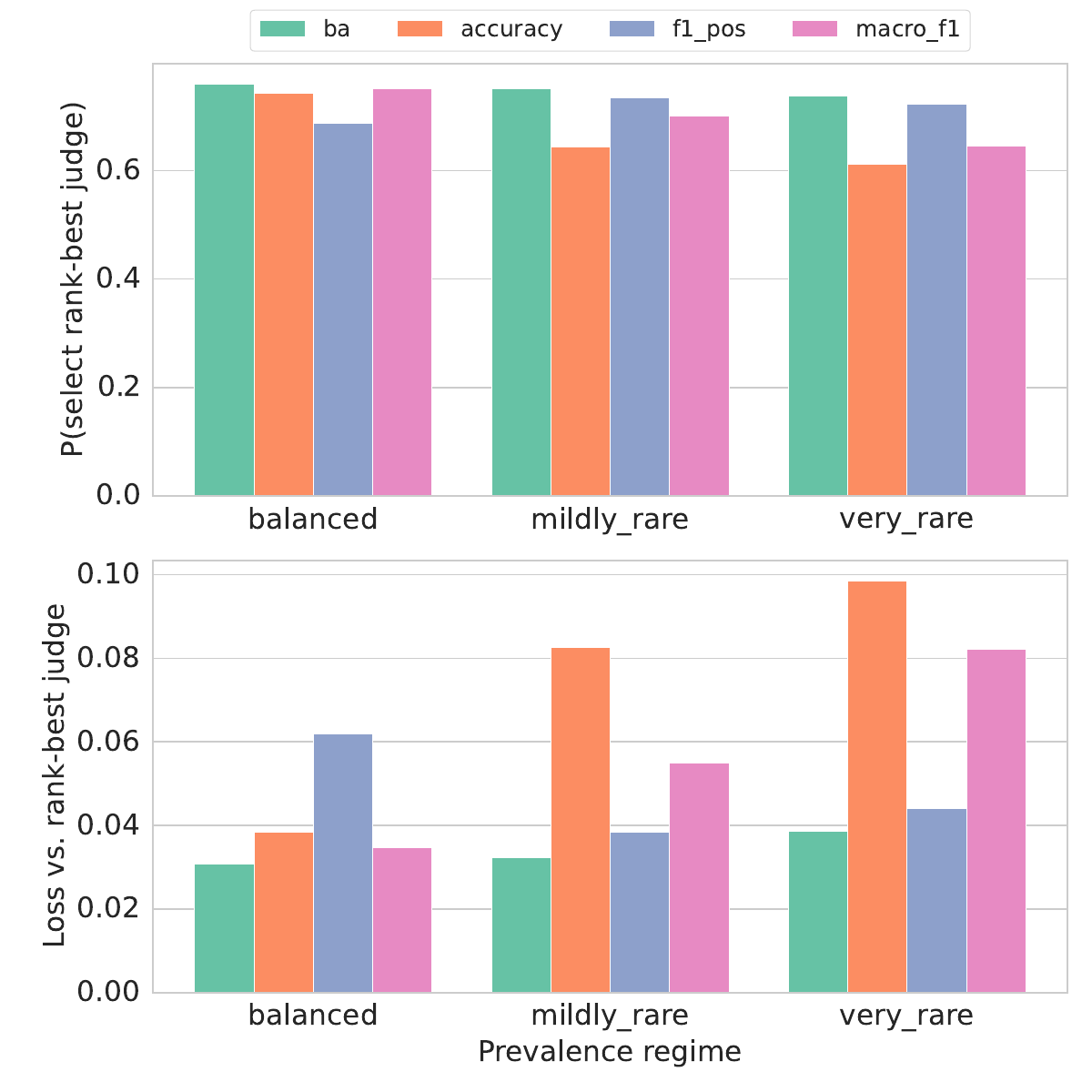}
    \caption{Balanced Accuracy is more robust to imbalance compared to other selection metrics across different prevalence regimes. Its ranking loss increases only slightly as the behavior becomes rarer.}
    \label{fig:simulation_1}
\end{figure}

Balanced Accuracy not only picks the rank-optimal judge more often than any competing metric, but when it does mis-select, the resulting judge is still much closer to optimal. Relative to Accuracy, using Balanced Accuracy instead cuts the ranking-accuracy loss by roughly 51\% (from 0.067 to 0.033); relative to Macro-F1 and F1, the reduction is around 33–65\%.

\paragraph{Effect of golden-set prevalence.}

We next vary the class prevalence in the golden set, sampling $p_{\text{golden}}$ from three regimes:

\begin{itemize}
    \item Balanced: $p_{\text{golden}} \in [0.3, 0.7]$
    \item Mildly rare: $p_{\text{golden}} \in [0.05, 0.2]$
    \item Very rare: $p_{\text{golden}} \in [0.005, 0.05]$
\end{itemize}

Balanced Accuracy is consistently the best or tied-best metric in terms of selection probability, and it always yields the smallest average ranking loss (Figure \ref{fig:simulation_1}).

By contrast, Accuracy's ranking loss more than doubles from the balanced to very-rare regime, and it becomes the worst-performing metric in both selection probability and ranking loss. The more imbalanced or rare the behavior, the more one “pays” for using Accuracy (or Macro-F1) instead of Balanced Accuracy as the judge-selection criterion.

\paragraph{Effect of golden-set size.}

We vary the size of the golden set, $n_{\text{golden}}$ from 25 to 51{,}000 samples, while holding $n_{\text{eval}}$ fixed.

We observe diminishing returns in $n_{\text{golden}}$. Going from 25 to a few thousand labeled examples reduces ranking loss for all metrics, but beyond about 1{,}000--2{,}000 labels the curves largely plateau.

\begin{figure}[t]
    \centering
    \includegraphics[width=\linewidth]{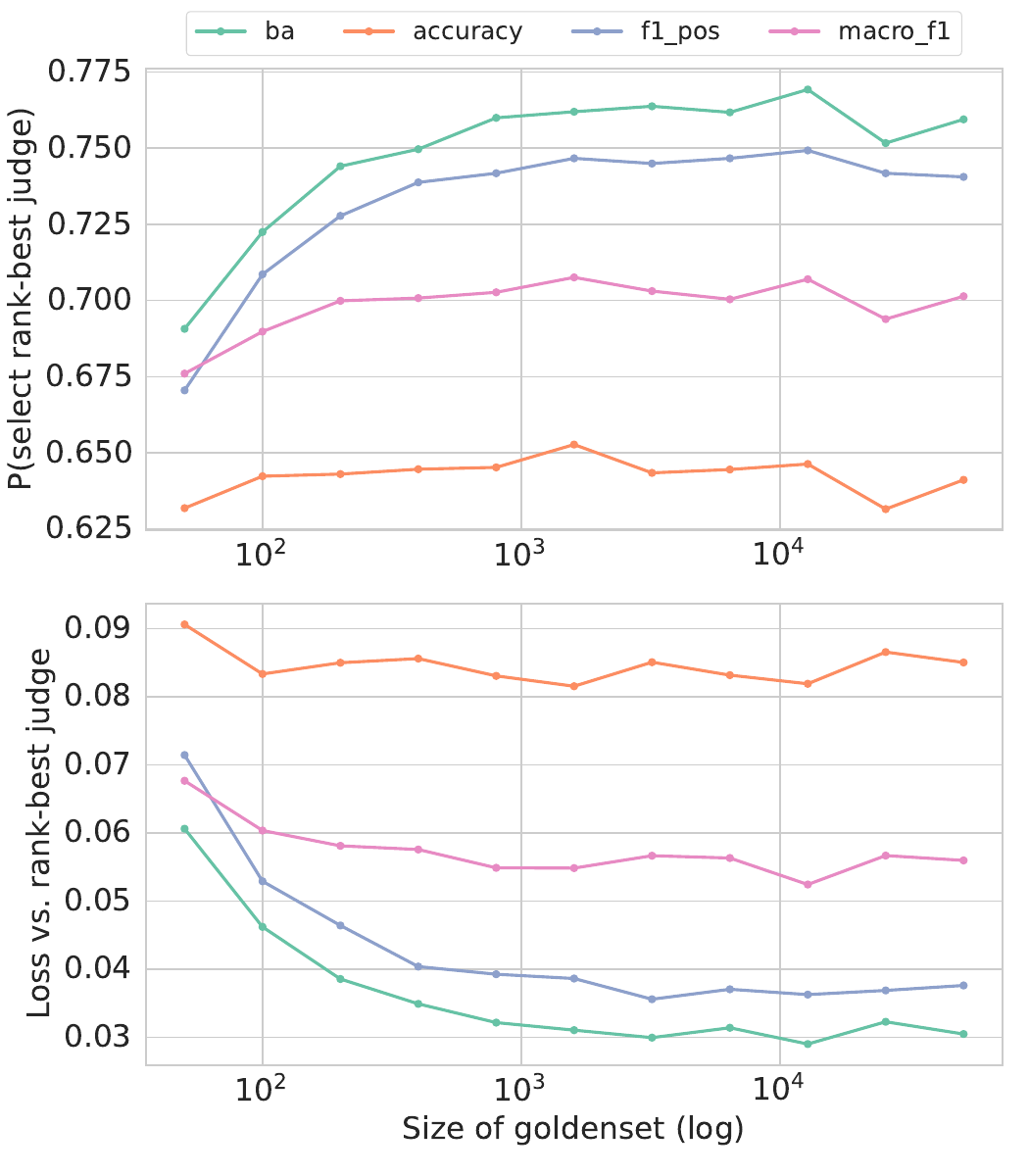}
    \caption{Going from 25 to a few thousand labeled examples reduces ranking loss for all metrics, but beyond about 1{,}000--2{,}000 labels the curves largely plateau. Balanced Accuracy is the clear winning metric for selecting a rank-optimal judge.}
    \label{fig:simulation_2}
\end{figure}

Balanced Accuracy dominates for all $n_{\text{golden}}$. It consistently has the highest probability of selecting the rank-optimal judge and the smallest average RankAcc gap, with a particularly large margin over Accuracy. At large golden-set sizes, BA’s ranking loss is less than half that of Accuracy (Figure \ref{fig:simulation_2}).

Collecting a moderately sized golden set is helpful, but once past the low-data regime, metric choice becomes more important than squeezing out a few extra labels.

\paragraph{Effect of model-evaluation sample size.}

Finally, we vary the number of model-eval samples per assistant model, $n_{\text{eval}}$ from 25 to 51{,}000 samples, while holding the golden-set size fixed.

\begin{figure}[t]
    \centering
    \includegraphics[width=\linewidth]{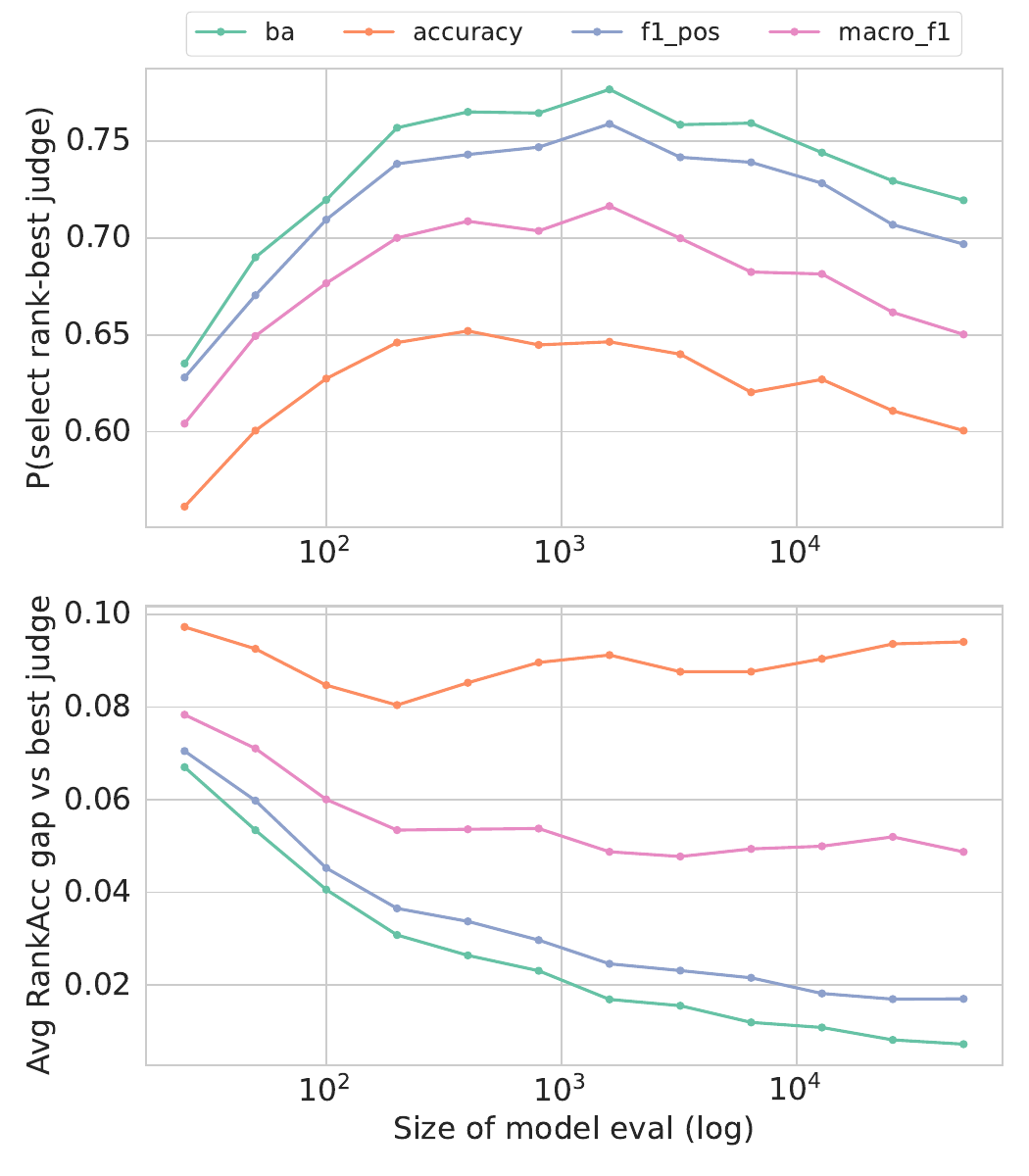}
    \caption{Across all model evaluation sample sizes, Balanced Accuracy's selection probability curve is the highest, typically sitting 2–5 percentage points above F1/Macro-F1 and 10+ points above Accuracy at moderate to large $n_{\text{eval}}$.}
    \label{fig:simulation_3}
\end{figure}

As $n_{\text{eval}}$ increases, the RankAcc of each judge is estimated more accurately, and all metrics benefit: the average RankAcc gap between the metric-selected judge and the rank-optimal judge drops monotonically. Across all sample sizes, however, Balanced Accuracy is uniformly best (Figure \ref{fig:simulation_3}).

\end{document}